\documentclass[10pt,twocolumn,letterpaper]{article}

\usepackage{iccv}
\usepackage{times}
\usepackage{epsfig}
\usepackage{graphicx}
\usepackage{amsmath}
\usepackage{amssymb}
\usepackage{algorithm}
\usepackage{algorithmicx}
\usepackage{algpseudocode}
\usepackage{spverbatim}
\usepackage{etoolbox}
\makeatletter
\preto{\@verbatim}{\topsep=1pt \partopsep=1pt}
\makeatother

\usepackage{booktabs}
\usepackage[x11names,dvipsnames,table]{xcolor} 
\usepackage{multirow}
\usepackage{array}
\newcolumntype{L}[1]{>{\raggedright\let\newline\\\arraybackslash\hspace{0pt}}m{#1}}
\newcolumntype{C}[1]{>{\centering\let\newline\\\arraybackslash\hspace{0pt}}m{#1}}
\newcolumntype{R}[1]{>{\raggedleft\let\newline\\\arraybackslash\hspace{0pt}}m{#1}}

\usepackage[breaklinks=true,bookmarks=false]{hyperref}

\iccvfinalcopy 

\def\iccvPaperID{****} 

\begin{document}

\title{Genetic CNN}

\author{Lingxi Xie, Alan Yuille \\
Center for Imaging Science, The Johns Hopkins University, Baltimore, MD, USA \\
{\tt\small 198808xc@gmail.com}\quad{\tt\small alan.l.yuille@gmail.com}
}

\maketitle

\begin{abstract}
The deep Convolutional Neural Network (CNN) is the state-of-the-art solution for large-scale visual recognition.
Following basic principles such as increasing the depth and constructing highway connections,
researchers have manually designed a lot of fixed network structures and verified their effectiveness.

In this paper, we discuss the possibility of learning deep network structures automatically.
Note that the number of possible network structures increases exponentially with the number of layers in the network,
which inspires us to adopt the genetic algorithm to efficiently traverse this large search space.
We first propose an encoding method to represent each network structure in a fixed-length binary string,
and initialize the genetic algorithm by generating a set of randomized individuals.
In each generation, we define standard genetic operations, e.g., selection, mutation and crossover,
to eliminate weak individuals and then generate more competitive ones.
The competitiveness of each individual is defined as its recognition accuracy,
which is obtained via training the network from scratch and evaluating it on a validation set.
We run the genetic process on two small datasets, i.e., {\bf MNIST} and {\bf CIFAR10},
demonstrating its ability to evolve and find high-quality structures which are little studied before.
These structures are also transferrable to the large-scale {\bf ILSVRC2012} dataset.
\end{abstract}

\section{Introduction}
\label{Introduction}

Visual recognition is a fundamental task in computer vision, implying a wide range of applications.
Recently, the state-of-the-art algorithms on visual recognition are mostly based on the deep Convolutional Neural Network (CNN).
Starting from the fundamental network model for large-scale image classification~\cite{Krizhevsky_2012_ImageNet},
researchers have been increasing the depth of the network~\cite{Simonyan_2015_Very},
as well as designing new inner structures~\cite{Szegedy_2015_Going}\cite{He_2016_Deep} to improve recognition accuracy.
Although these modern networks have been shown to be efficient,
we note that their structures are manually designed, not learned, which limits the flexibility of the approach.

In this paper, we explore the possibility of automatically learning the structure of deep neural networks.
We consider a constrained case, in which the network has a limited number of layers,
and each layer is designed as a set of pre-defined building blocks such as convolution and pooling.
Even under these limitations, the total number of possible network structures grows exponentially with the number of layers.
Therefore, it is impractical to enumerate all the candidates and find the best one.
Instead, we formulate this problem as optimization in a large search space,
and apply the genetic algorithm to traversing the space efficiently.

The genetic algorithm involves constructing an initial {\em generation} of {\em individuals} (candidate solutions),
and performing genetic operations to allow them to evolve in a genetic process.
To this end, we propose an encoding method to represent each network structure by a fixed-length binary string.
After that, we define several standard genetic operations, {\em i.e.}, selection, mutation and crossover,
which eliminate weak individuals of the previous generation and use them to generate competitive ones.
The quality of each individual is determined by its recognition accuracy on a reference dataset.
{\bf Throughout the genetic process, we evaluate each individual ({\em i.e.}, network structure) by training it from scratch.}
The genetic process comes to an end after a fixed number of generations.

It is worth emphasizing that the genetic algorithm is computationally expensive,
because we need to conduct a complete network training process for each generated individual.
Therefore, we run the genetic process on two small datasets, {\em i.e.}, {\bf MNIST} and {\bf CIFAR10},
and demonstrate its ability to find high-quality network structures.
{\bf It is interesting to see that the generated structures, most of which have been less studied before,
often perform better than the standard manually designed ones.}
Finally, we transfer the learned structures to large-scale experiments and verify their effectiveness.

The remainder of this paper is organized as follows.
Section~\ref{RelatedWork} briefly introduces related work.
Section~\ref{Algorithm} illustrates the way of using the genetic algorithm to design network structures.
Experiments are shown in Section~\ref{Experiments}, and conclusions are drawn in Section~\ref{Conclusions}.

\section{Related Work}
\label{RelatedWork}

\subsection{Convolutional Neural Networks}
\label{RelatedWork:CNN}

Image classification is a fundamental problem in computer vision.
Recently, researcher have extended conventional classification tasks~\cite{Lazebnik_2006_Beyond}\cite{Feifei_2007_Learning}
into large-scale environments such as ImageNet~\cite{Deng_2009_ImageNet} and Places~\cite{Zhou_2014_Learning}.
With the availability of powerful computational resources ({\em e.g.}, GPU),
the Convolutional Neural Networks (CNNs)~\cite{Krizhevsky_2012_ImageNet}\cite{Simonyan_2015_Very} have shown superior performance
over the conventional Bag-of-Visual-Words models~\cite{Csurka_2004_Visual}\cite{Wang_2010_Locality}\cite{Perronnin_2010_Improving}.

CNN is a hierarchical model for large-scale visual recognition.
It is based on the observation that a network with enough neurons is able to fit any complicated data distribution.
In past years, neural networks were shown effective for simple recognition tasks~\cite{LeCun_1990_Handwritten}.
More recently, the availability of large-scale training data ({\em e.g.}, ImageNet~\cite{Deng_2009_ImageNet}) and powerful GPUs
make it possible to train deep CNNs~\cite{Krizhevsky_2012_ImageNet} which significantly outperform BoVW models.
A CNN is composed of several stacked layers.
In each of them, responses from the previous layer are convoluted with a filter bank and activated by a differentiable non-linearity.
Hence, a CNN can be considered as a composite function,
which is trained by back-propagating error signals defined by the difference between the supervision and prediction at the top layer.
Recently, several efficient methods were proposed to help CNNs converge faster and prevent over-fitting,
such as ReLU activation~\cite{Krizhevsky_2012_ImageNet}, batch normalization~\cite{Ioffe_2015_Batch},
Dropout~\cite{Hinton_2012_Improving} and DisturbLabel~\cite{Xie_2016_DisturbLabel}.
Features extracted from pre-trained neural networks can be generalized to
other recognition tasks~\cite{Xie_2015_Image}\cite{Xie_2016_InterActive}.

Designing powerful CNN structures is an intriguing problem.
It is believed that deeper networks produce better recognition results~\cite{Simonyan_2015_Very}\cite{Szegedy_2015_Going}.
But also, adding highway information has been verified to be useful~\cite{He_2016_Deep}\cite{Zagoruyko_2016_Wide}.
Efforts are also made to add invariance into the network structure~\cite{Xie_2016_Towards}.
We find some work which uses stochastic~\cite{Huang_2016_Deep} or dense~\cite{Huang_2016_Densely} structures,
but all these network structures are deterministic
(although stochastic operations are used in~\cite{Huang_2016_Deep} to accelerate training and prevent over-fitting),
which limits the flexibility of the models and thus inspires us to automatically learn network structures.

\subsection{Genetic Algorithm}
\label{RelatedWork:GeneticAlgorithm}

The genetic algorithm is a metaheuristic inspired by the process of natural selection.
Genetic algorithms are commonly used to generate high-quality solutions to optimization and search
problems~\cite{Houck_1995_Genetic}\cite{Reeves_1995_Genetic}\cite{Beasley_1996_Genetic}\cite{Deb_2002_Fast}
by relying on bio-inspired operators such as mutation, crossover and selection.

A typical genetic algorithm requires two prerequisites,
{\em i.e.}, a genetic representation of the solution domain, and a fitness function to evaluate each individual.
A good example is the travelling-salesman problem (TSP)~\cite{Grefenstette_1985_Genetic},
a famous NP-complete problem which aims at finding the optimal Hamiltonian path in a graph of $N$ nodes.
In this situation, each feasible solution is represented as a permutation of $\left\{1,2,\ldots,N\right\}$,
and the fitness function is the total cost (distance) of the path.
We will show later that deep neural networks can be encoded into a binary string.

The core idea of the genetic algorithm is to allow individuals to evolve via some genetic operations.
Popular operations include {\em selection}, {\em mutation}, {\em crossover}, {\em etc}.
The selection process allows us to preserve strong individuals while eliminating weak ones.
The ways of performing mutation and crossover vary from case to case, often based on the properties of the specific problem.
For example, in the TSP problem with the permutation-based representation,
a possible set of mutations is to change the order of two visited nodes.
These operations are also used in our work.

There is a lot of research in how to improve the performance of genetic algorithms,
including performing local search~\cite{Ulder_1990_Genetic} and generating random keys~\cite{Snyder_2006_Random}.
In our work, we show that the vanilla genetic algorithm works well enough without these tricks.
We also note that some previous work applied the genetic algorithm to exploring efficient neural network
architectures~\cite{Yao_1999_Evolving}\cite{Stanley_2002_Evolving}\cite{Bayer_2009_Evolving}\cite{Ding_2013_Evolutionary},
but our work aims at learning the architecture of modern CNNs, which is not studied in these prior works.

\section{Our Approach}
\label{Algorithm}

This section presents a genetic algorithm for designing competitive network structures.
First, we describe a way of representing the network structure by a fixed-length binary string.
Next, several genetic operations are defined, including selection, mutation and crossover,
so that we can traverse the search space efficiently and find high-quality solutions.

{\bf Throughout this work, the genetic algorithm is only used to propose new network structures,
the parameters and classification accuracy of each structure are obtained via standalone training-from-scratch.}

\subsection{Binary Network Representation}
\label{Algorithm:Representation}

\newcommand{\bigfigurewidth}{14.0cm}
\begin{figure*}
\begin{center}
    \includegraphics[width=\bigfigurewidth]{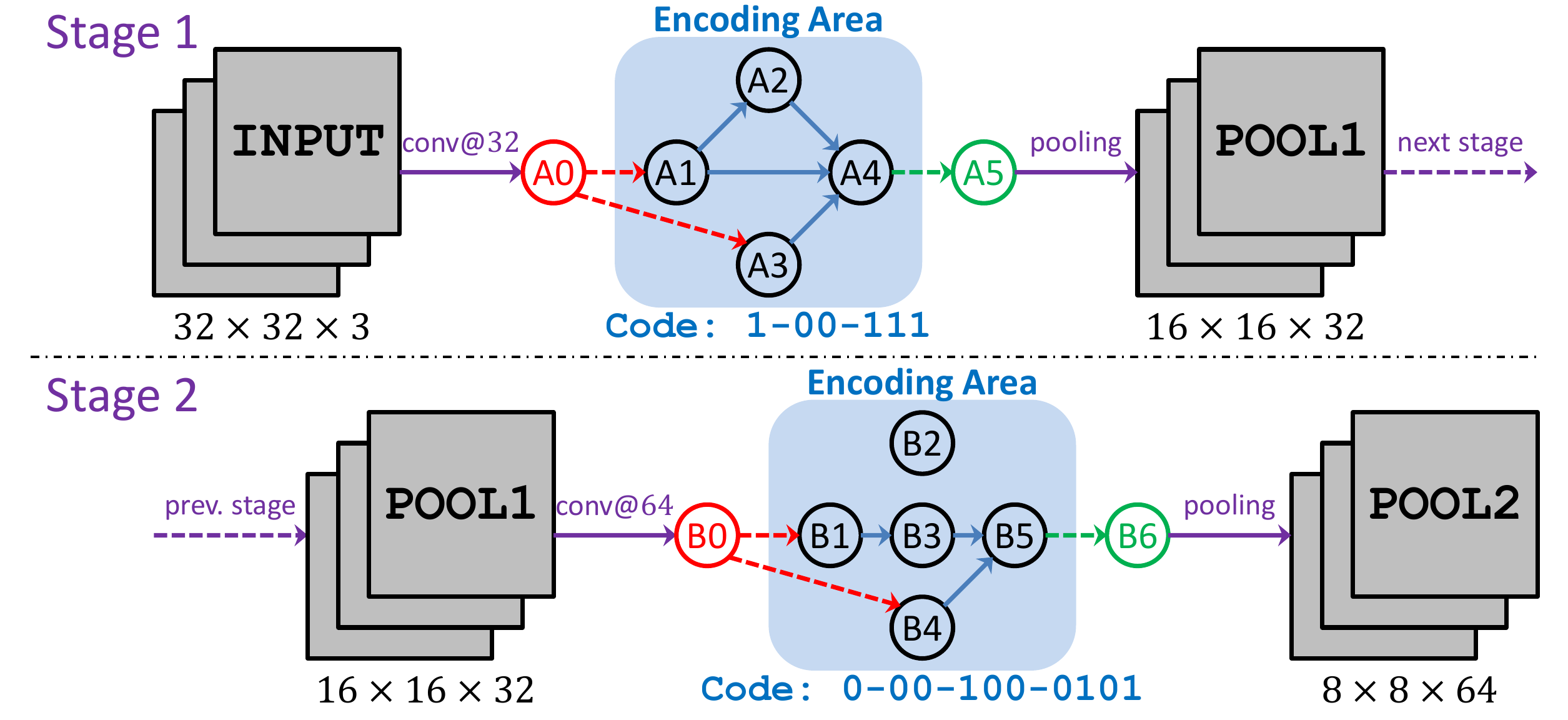}
\end{center}
\caption{
    A two-stage network (${S}={2}$, ${\left(K_1,K_2\right)}={\left(4,5\right)}$)
    and the encoded binary string (best viewed in color PDF).
    The default input and output nodes (see Section~\ref{Algorithm:Representation:Details})
    and the connections from and to these nodes are marked in red and green, respectively.
    We only encode the connections in the effective parts (regions with light blue background).
    Within each stage, the number of convolutional filters is a constant ($32$ in Stage $1$, $64$ in Stage $2$),
    and the spatial resolution remains unchanged ($32\times32$ in Stage $1$, $16\times16$ in Stage $2$).
    Each pooling layer down-samples the data by a factor of $2$.
    ReLU and batch normalization are added after each convolution.
}
\label{Fig:NetworkEncoding}
\end{figure*}

We provide a binary string representation for a network structure in a constrained case.
We first note that many state-of-the-art network structures~\cite{Simonyan_2015_Very}\cite{He_2016_Deep}
can be partitioned into several {\em stages}.
In each stage, the geometric dimensions (width, height and depth) of the layer cube remain unchanged.
Neighboring stages are connected via a spatial pooling operation, which may change the spatial resolution.
All the convolutional operations within one stage have the same number of filters, or channels.

We borrow this idea to define a family of networks which can be encoded into fixed-length binary strings.
A network is composed of $S$ stages, and the $s$-th stage, ${s}={1,2,\ldots,S}$, contains $K_s$ {\em nodes},
denoted by $v_{s,k_s}$, ${k_s}={1,2,\ldots,K_s}$.
The nodes within each stage are ordered, and we only allow connections from a lower-numbered node to a higher-numbered node.
Each node corresponds to a convolutional operation,
which takes place after element-wise summing up all its input nodes (lower-numbered nodes that are connected to it).
After convolution, batch normalization~\cite{Ioffe_2015_Batch} and ReLU~\cite{Krizhevsky_2012_ImageNet} are followed,
which are verified efficient in training very deep neural networks~\cite{Simonyan_2015_Very}.
We do not encode the fully-connected part of a network.

In each stage, we use ${1+2+\ldots+\left(K_s-1\right)}={\frac{1}{2}K_s\left(K_s-1\right)}$ bits to encode the inter-node connections.
The first bit represents the connection between $\left(v_{s,1},v_{s,2}\right)$,
then the following two bits represent the connection
between $\left(v_{s,1},v_{s,3}\right)$ and $\left(v_{s,2},v_{s,3}\right)$, {\em etc}.
This process continues until the last $K_s-1$ bits are used to represent the connection
between $v_{s,1},v_{s,2},\ldots,v_{s,K_s-1}$ and $v_{s,K_s}$.
For ${1}\leqslant{i}<{j}\leqslant{K_s}$,
if the code corresponding to $\left(v_{s,i},v_{s,j}\right)$ is $1$, there is an edge connecting $v_{s,i}$ and $v_{s,j}$,
{\em i.e.}, $v_{s,j}$ takes the output of $v_{s,i}$ as a part of the element-wise summation, and vice versa.

Figure~\ref{Fig:NetworkEncoding} illustrates two examples of network encoding.
To summarize, a $S$-stage network with $K_s$ nodes at the $s$-th stage
is encoded into a binary string with length ${L}={\frac{1}{2}{\sum_s}K_s\left(K_s-1\right)}$.
Equivalently, there are in total $2^L$ possible network structures.
This number may be very large.
In the {\bf CIFAR10} experiments (see Section~\ref{Experiments:CIFAR10}),
we have ${S}={3}$ and ${\left(K_1,K_2,K_3\right)}={\left(3,4,5\right)}$, therefore ${L}={19}$ and ${2^L}={524\rm{,}288}$.
It is computationally intractable to enumerate all these structures and find the optimal one(s).
In the following parts, we adopt the genetic algorithm to efficiently explore good candidates in this large space.

\subsubsection{Technical Details}
\label{Algorithm:Representation:Details}

To make every binary string valid, we define two default nodes in each stage.
The default input node, denoted as $v_{s,0}$, receives data from the previous stage, performs convolution,
and sends its output to every node without a predecessor, {\em e.g.}, $v_{s,1}$.
The default output node, denoted as $v_{s,K_s+1}$, receives data from all nodes without a successor, {\em e.g.}, $v_{s,K_s}$,
sums up them, performs convolution, and sends its output to the pooling layer.
Note that the connections between the ordinary nodes and the default nodes are not encoded.

There are two special cases.
First, if an ordinary node $v_{s,i}$ is isolated
({\em i.e.}, it is not connected to any other ordinary nodes $v_{s,j}$, ${i}\neq{j}$),
then it is simply ignored, {\em i.e.}, it is not connected to the default input node nor the default output node
(see the {\tt B2} node in Figure~\ref{Fig:NetworkEncoding}).
This is to guarantee that a stage with more nodes can simulate all structures represented by a stage with fewer nodes.
Second, if there are no connections at a stage, {\em i.e.}, all bits in the binary string are $0$,
then the convolutional operation is performed only once, not twice
(one for the default input node and one for the default output node).

\subsubsection{Examples and Limitations}
\label{Algorithm:Representation:Examples}

\newcommand{\figurewidth}{8.0cm}
\begin{figure}
\begin{center}
    \includegraphics[width=\figurewidth]{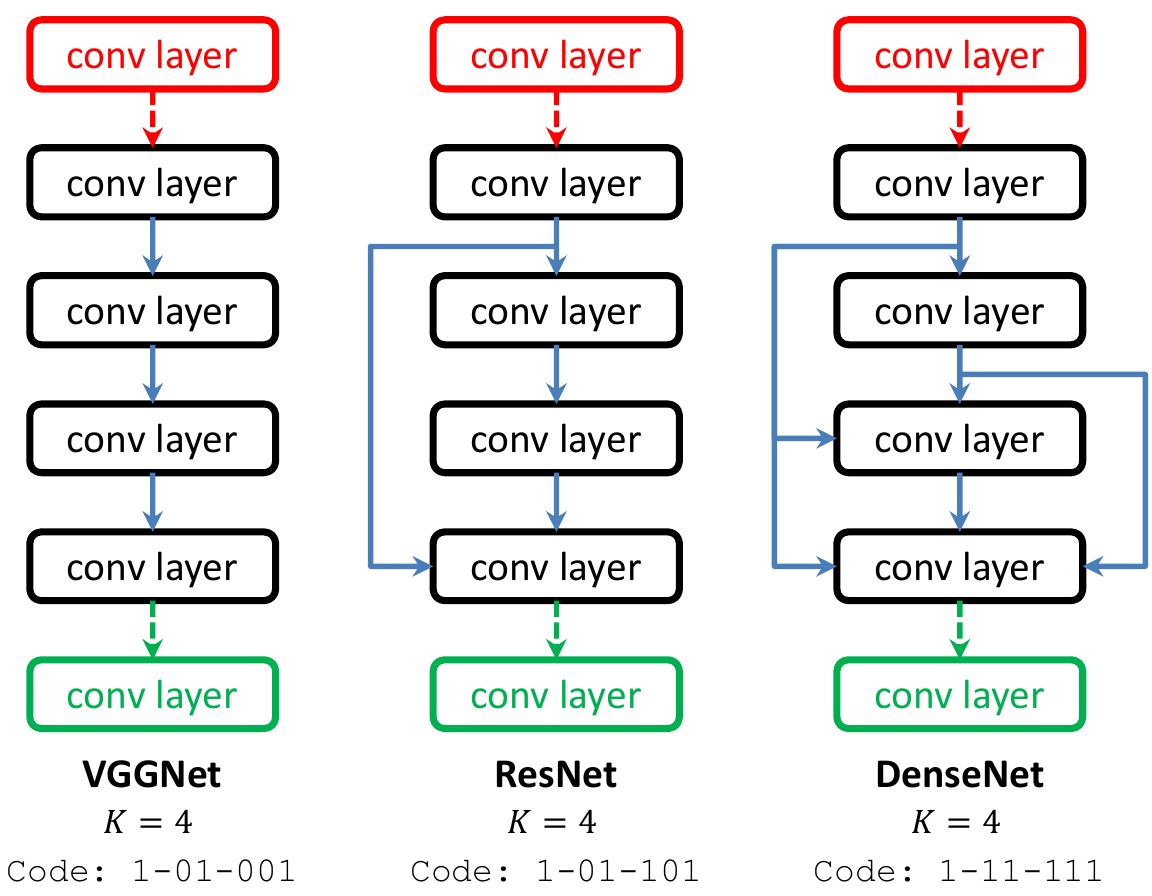}
\end{center}
\caption{
    The basic building blocks of {\bf VGGNet}~\cite{Simonyan_2015_Very} and {\bf ResNet}~\cite{He_2016_Deep}
    can be encoded as binary strings defined in Section~\ref{Algorithm:Representation}.
}
\label{Fig:Examples}
\end{figure}

Many popular network structures can be represented using the proposed encoding scheme.
Examples include {\bf VGGNet}~\cite{Simonyan_2015_Very}, {\bf ResNet}~\cite{He_2016_Deep},
and a modified variant of {\bf DenseNet}~\cite{Huang_2016_Densely},
which are illustrated in Figure~\ref{Fig:Examples}.

Currently, only convolutional and pooling operations are considered,
which makes it impossible to generate some tricky network modules such as Maxout~\cite{Goodfellow_2013_Maxout}.
Also, the size of convolutional filters is fixed within each stage,
which limits our network from incorporating multi-scale information as in the inception module~\cite{Szegedy_2015_Going}.
However, we note that all the encoding-based approaches have such limitations.
Our approach can be easily modified to include more types of layers and more flexible inter-layer connections.
As shown in experiments, we can achieve competitive recognition performance using merely these basic building blocks.

As shown in a recent published work using reinforcement learning to explore neural architecture~\cite{Zoph_2017_Neural},
this type of methods often require heavy computation to traverse the huge solution space.
Fortunately, our method can be easily generalized and scaled up,
which is done via learning the architecture on a small dataset and transfer the learned information to large-scale datasets.
Please refer to the experimental part for details.

\subsection{Genetic Operations}
\label{Algorithm:Operations}

\begin{algorithm*}
\caption{The Genetic Process for Network Design}
\begin{algorithmic}[1]
\State {\bf Input:} the reference dataset $\mathcal{D}$,
    the number of generations $T$, the number of individuals in each generation $N$,
    the mutation and crossover probabilities $p_\mathrm{M}$ and $p_\mathrm{C}$,
    the mutation parameter $q_\mathrm{M}$, and the crossover parameter $q_\mathrm{C}$.
\State {\bf Initialization:} generating a set of randomized individuals $\left\{\mathbb{M}_{0,n}\right\}_{n=1}^N$,
    and computing their recognition accuracies;
\For {${t}={1,2,\ldots,T}$}{}
\State {\bf Selection:} producing a new generation $\left\{\mathbb{M}'_{t,n}\right\}_{n=1}^N$
    with a Russian roulette process on $\left\{\mathbb{M}_{t-1,n}\right\}_{n=1}^N$;
\State {\bf Crossover:} for each pair
    $\left\{\left(\mathbb{M}_{t,2n-1},\mathbb{M}_{t,2n}\right)\right\}_{n=1}^{\left\lfloor N/2\right\rfloor}$,
    performing crossover with probability $p_\mathrm{C}$ and parameter $q_\mathrm{C}$;
\State {\bf Mutation:} for each non-crossover individual $\left\{\mathbb{M}_{t,n}\right\}_{n=1}^N$,
    doing mutation with probability $p_\mathrm{M}$ and parameter $q_\mathrm{M}$;
\State {\bf Evaluation:} computing the recognition accuracy for each new individual $\left\{\mathbb{M}_{t,n}\right\}_{n=1}^N$;
\EndFor
\State {\bf Output:} a set of individuals in the final generation $\left\{\mathbb{M}_{T,n}\right\}_{n=1}^N$
    with their recognition accuracies.
\end{algorithmic}
\label{Alg:GeneticAlgorithm}
\end{algorithm*}

The flowchart of the genetic process is shown in Algorithm~\ref{Alg:GeneticAlgorithm}.
It starts with an initialized {\em generation} of $N$ randomized {\em individuals}.
Then, we perform $T$ rounds, or $T$ generations,
each of which consists of three operations, {\em i.e.}, selection, mutation and crossover.
The fitness function of each individual is evaluated via training-from-scratch on the reference dataset.

\subsubsection{Initialization}
\label{Algorithm:Operations:Initialization}

We initialize a set of randomized models $\left\{\mathbb{M}_{0,n}\right\}_{n=1}^N$.
Each model is a binary string with $L$ bits, {\em i.e.}, ${\mathbb{M}_{0,n}}:{\mathbf{b}_{0,n}}\in{\left\{0,1\right\}^L}$.
Each bit in each individual is independently sampled from a Bernoulli distribution:
${b_{0,n}^{l}}\sim{\mathcal{B}\!\left(0.5\right)}$, ${l}={1,2,\ldots,L}$.
After this, we evaluate each individual (see Section~\ref{Algorithm:Operations:Evaluation}) to obtain their fitness function values.

As we shall see in Section~\ref{Experiments:MNIST:Initialization},
different strategies of initialization do not impact the genetic performance too much.
Even starting with a naive initialization (all individuals are all-zero strings),
the genetic process can discover quite competitive structures with crossover and mutation.

\subsubsection{Selection}
\label{Algorithm:Operations:Selection}

The selection process is performed at the beginning of every generation.
Before the $t$-th generation, the $n$-th individual $\mathbb{M}_{t-1,n}$ is assigned a fitness function,
which is defined as the recognition rate $r_{t-1,n}$ obtained in the previous generation or initialization.
$r_{t-1,n}$ directly impacts the probability that $\mathbb{M}_{t-1,n}$ survives the selection process.

We perform a Russian roulette process to determine which individuals survive.
Each individual in the next generation $\mathbb{M}_{t,n}$ is determined independently
by a non-uniform sampling over the set $\left\{\mathbb{M}_{t-1,n}\right\}_{n=1}^N$.
The probability of sampling $\mathbb{M}_{t-1,n}$ is proportional to $r_{t-1,n}-r_{t-1,0}$,
where ${r_{t-1,0}}={{\min_{n=1}^N}\left\{r_{t-1,n}\right\}}$ is the minimal fitness function value in the previous generation.
This means that the best individual has the largest probability of being selected, and the worst one is always eliminated.
As the number of individuals $N$ remains unchanged, each individual in the previous generation may be selected multiple times.

\subsubsection{Mutation and Crossover}
\label{Algorithm:Operations:MutationCrossover}

The mutation process of an individual $\mathbb{M}_{t,n}$ involves flipping each bit independently with a probability $q_\mathrm{M}$.
In practice, $q_\mathrm{M}$ is often small, {\em e.g.}, $0.05$, so that mutation is not likely to change one individual too much.
This is to preserve the good properties of a survived individual while providing an opportunity of trying out new possibilities.

The crossover process involves changing two individuals simultaneously.
Instead of considering each bit individually, the basic unit in crossover is a stage,
which is motivated by the need to retain the local structures within each stage.
Similar to mutation, each pair of corresponding stages are exchanged with a small probability $q_\mathrm{C}$.

Both mutation and crossover are implemented by an overall flowchart (see Algorithm~\ref{Alg:GeneticAlgorithm}).
The probabilities of mutation and crossover for each individual (or pair) are $p_\mathrm{M}$ and $p_\mathrm{C}$, respectively.
We understand that there are many different ways of mutation and crossover.
As shown in experiments, our simple choice leads to competitive performance.

\subsubsection{Evaluation}
\label{Algorithm:Operations:Evaluation}

After the above processes, each individual $\mathbb{M}_{t,n}$ is evaluated to obtain the fitness function value.
A reference dataset $\mathcal{D}$ is pre-defined, and we individually train each model $\mathbb{M}_{t,n}$ from scratch.
If $\mathbb{M}_{t,n}$ is previously evaluated,
we simply evaluate it once again and compute the average accuracy over all its occurrences.
This strategy, at least to some extent, alleviates the instability caused by the randomness in the training process.

\section{Experiments}
\label{Experiments}

The proposed genetic algorithm requires a very large amount of computational resources,
which makes it intractable to be directly evaluated on large-scale datasets such as {\bf ILSVRC2012}~\cite{Russakovsky_2015_ImageNet}.
Our solution is to explore promising network structures
on small datasets such as {\bf MNIST}~\cite{LeCun_1998_Gradient} and {\bf CIFAR10}~\cite{Krizhevsky_2009_Learning},
then transfer these structures to the large-scale recognition tasks.

\subsection{MNIST Experiments}
\label{Experiments:MNIST}

The {\bf MNIST} dataset~\cite{LeCun_1998_Gradient} defines a handwritten digit recognition task.
There are $60\rm{,}000$ images for training, and $10\rm{,}000$ images for testing, all of them are $28\times28$ grayscale images.
Both training and testing data are uniformly distributed over $10$ categories, {\em i.e.}, digits from $0$ to $9$.
To avoid using the testing data, we leave $10\rm{,}000$ images from the training set for validation.

\subsubsection{Settings and Results}
\label{Experiments:MNIST:Results}

We follow the basic {\bf LeNet} for {\bf MNIST} recognition.
The original network is abbreviated as:
\begin{spverbatim}
C5@20-MP2S2-C5@50-MP2S2-FC500-D0.5-FC10.
\end{spverbatim}
\noindent
Here, {\tt C5@20} is a convolutional layer with a kernel size $5$, a default spatial stride $1$ and the number of kernels $20$;
{\tt MP2S2} is a max-pooling layer with a kernel size $2$ and a spatial stride $2$,
{\tt FC500} is a fully-connected layer with $500$ outputs, and {\tt D0.5} is a Dropout layer with a drop ratio $0.5$.
We apply $20$ training epochs with learning rate $10^{-3}$, followed by $4$ epochs with learning rate $10^{-4}$,
and another $1$ epoch with learning rate $10^{-5}$.

We set ${S}={2}$, ${\left(K_1,K_2\right)}={\left(3,5\right)}$, and keep the fully-connected part of {\bf LeNet} unchanged.
The first convolutional layer within each stage remains the same as in the original {\bf LeNet},
and other convolutional layers take the kernel size $3\times3$ and the same channel number.
The length $L$ of each binary string is $13$, which means that there are ${2^{13}={8\rm{,}192}}$ possible individuals.

We create an initial generation with ${N}={20}$ individuals, and run the genetic process for ${T}={50}$ rounds.
Other parameters are set as ${p_\mathrm{M}}={0.8}$, ${q_\mathrm{M}}={0.1}$, ${p_\mathrm{C}}={0.2}$ and ${q_\mathrm{C}}={0.3}$.
We set relatively high mutation and crossover probabilities to facilitate new structures to be generated.
The maximal number of explored individuals is ${20\times\left(50+1\right)}={1\rm{,}020}<{8\rm{,}192}$.
The training phase of each individual takes an average of $2.5$ minutes on a modern Titan-X GPU,
and the entire genetic process takes about $2$ GPU-days,
which makes it possible to repeat it with different settings for diagnosis,
{\em e.g.}, to explore different initialization options (see Section~\ref{Experiments:MNIST:Initialization}).

\begin{table}
\centering
\begin{tabular}{|l||r|r|r|r|r|}
\hline
Gen & Max $\%$         & Min $\%$         & Avg $\%$         & Med $\%$         & Std-D   \\
\hline
00  & $99.59$          & $99.38$          & $99.50$          & $99.50$          & $ 0.06$ \\
\hline
01  & $99.61$          & $99.40$          & $99.53$          & $99.54$          & $ 0.05$ \\
\hline
02  & $99.62$          & $99.43$          & $99.55$          & $99.58$          & $ 0.06$ \\
\hline
03  & $99.62$          & $99.40$          & $99.56$          & $99.58$          & $ 0.06$ \\
\hline
05  & $99.62$          & $99.46$          & $99.57$          & $99.57$          & $ 0.04$ \\
\hline
08  & $99.63$          & $99.40$          & $99.57$          & $99.60$          & $ 0.06$ \\
\hline
10  & $99.63$          & $99.50$          & $99.59$          & $99.62$          & $ 0.05$ \\
\hline
20  & $99.63$          & $99.45$          & $99.61$          & $99.63$          & $ 0.05$ \\
\hline
30  & $99.64$          & $99.49$          & $99.61$          & $99.64$          & $ 0.06$ \\
\hline
50  & $99.66$          & $99.51$          & $99.62$          & $99.65$          & $ 0.06$ \\
\hline
\end{tabular}
\caption{
    Recognition accuracy ($\%$) on the {\bf MNIST} testing set.
    The zeroth generation is the initialized generation.
    We set ${S}={2}$ and ${\left(K_1,K_2\right)}={\left(3,5\right)}$.
}
\label{Tab:MNIST}
\end{table}

Results are summarized in Table~\ref{Tab:MNIST}.
With the genetic operations, we can find competitive network structures which achieve high recognition accuracy.
Although over a short period the recognition rate of the best individual is not improved,
the average and medium accuracies generally get higher from generation to generation.
This is very important, because it guarantees the genetic algorithm improves the overall quality of the individuals.
According to our diagnosis in Section~\ref{Experiments:CIFAR10:Diagnosis}, this is very important for the genetic process,
since the quality of a new individual is positively correlated to the quality of its parent(s).
After $50$ generations, the recognition error rate of the best individual drops from $0.41\%$ to $0.34\%$.

\subsubsection{Diagnosis}
\label{Experiments:CIFAR10:Diagnosis}

\renewcommand{\figurewidth}{8.0cm}
\begin{figure}
\begin{center}
    \includegraphics[width=\figurewidth]{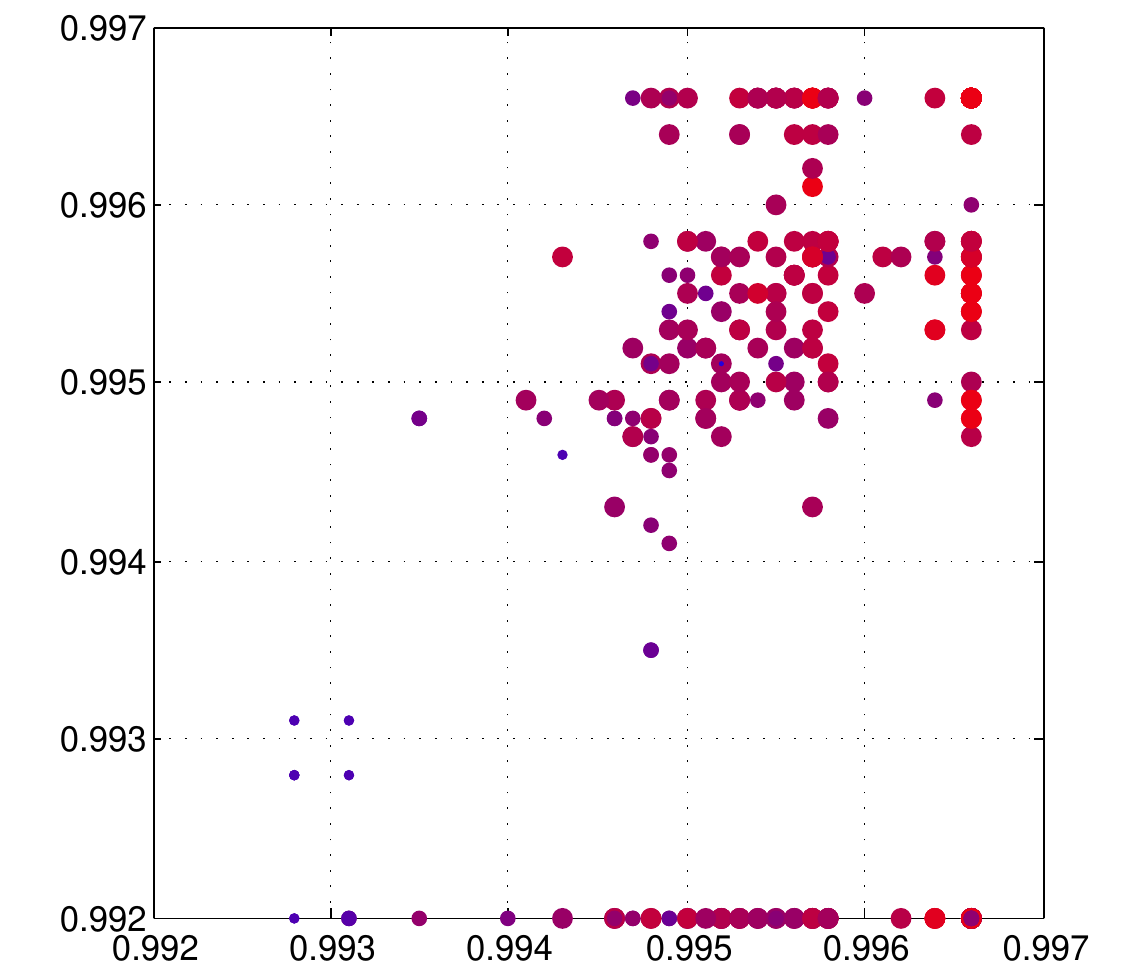}
\end{center}
\caption{
    The relationship in accuracy between the parent(s) and the child(ren) (best viewed in color PDF).
    A point is bigger and close to blue if the recognition error rate is lower, otherwise it is smaller and close to blue.
    The points on the horizontal axis are from mutation operations, while others are from crossover operations.
}
\label{Fig:Diagnosis}
\end{figure}

We perform diagnostic experiments to verify the hypothesis,
that a better individual is more likely to generate a good individual via mutation or crossover.
For this, we randomly select several occurrences of mutation and crossover in the {\bf CIFAR10} genetic process,
and observe the relationship between an individual and its parent(s).
Figure~\ref{Fig:Diagnosis} supports our point.
We argue that the genetic operations tend to preserve a fraction of the good local properties,
so that the excellent ``genes'' from the parent(s) are more likely to be preserved.

\subsubsection{Initialization Issues}
\label{Experiments:MNIST:Initialization}

\renewcommand{\figurewidth}{7.0cm}
\begin{figure}
\begin{center}
    \includegraphics[width=\figurewidth]{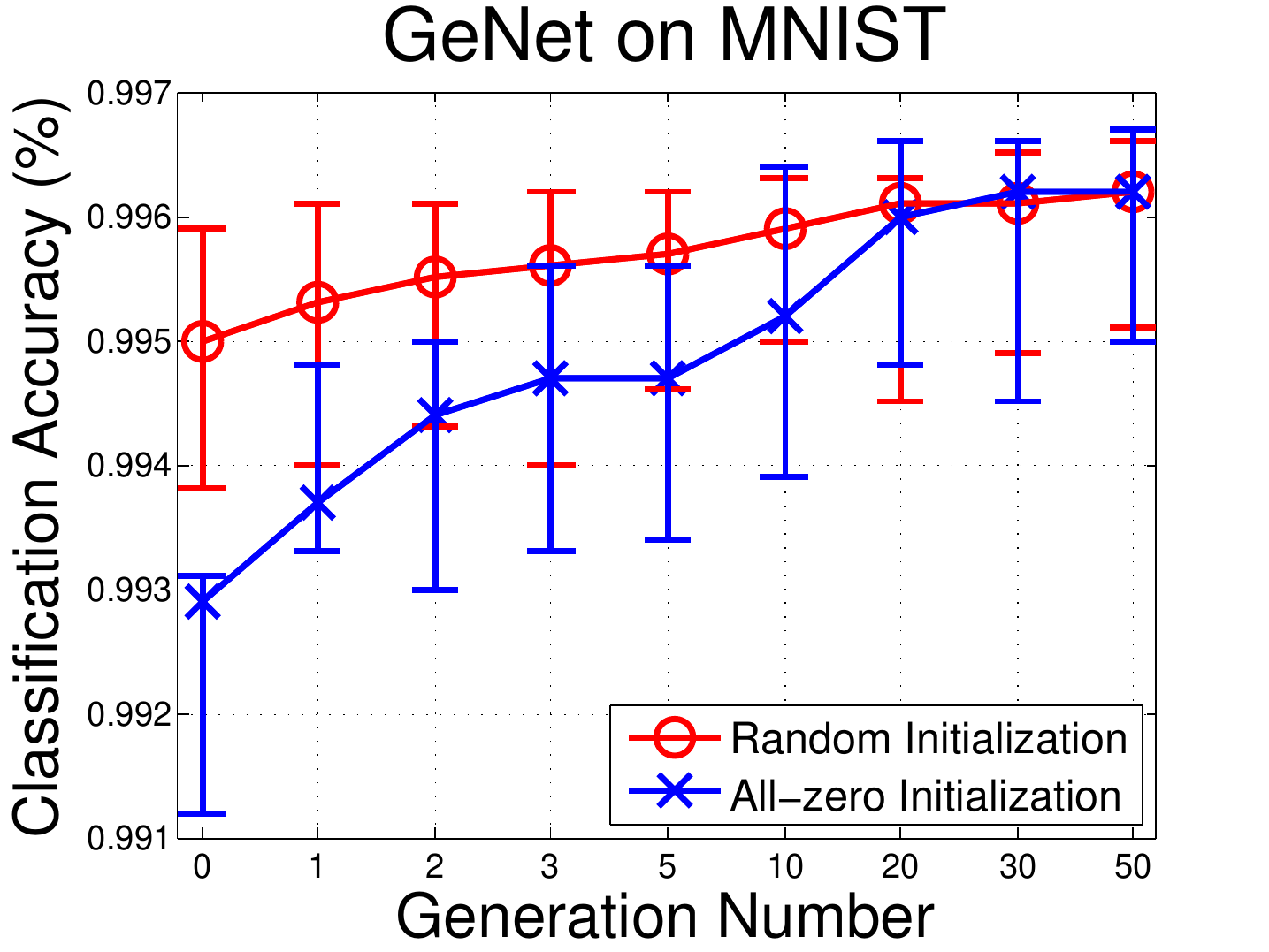}
\end{center}
\caption{
    The average recognition accuracy over all individuals with respect to the generation number.
    The bars indicate the highest and lowest accuracies in the corresponding generation.
}
\label{Fig:Initialization}
\end{figure}

Finally, we observe the impact of different initialized networks.
For this, we start a naive population with ${N}={20}$ all-zero individuals,
and use the same parameters for a complete genetic process.
Results are shown in Figure~\ref{Fig:Initialization}.
We find that, although the all-zero string corresponds to a very simple and less competitive network structure,
the genetic algorithm is able to generate strong individuals after several generations.
This naive initialization achieves the initial performance of randomized individuals with about $10$ generations.
After about $30$ generations, there is almost no difference, by statistics, between these two populations.

\subsection{CIFAR10 Experiments}
\label{Experiments:CIFAR10}

The {\bf CIFAR10} dataset~\cite{Krizhevsky_2009_Learning} is a subset of the $80$-million tiny image database~\cite{Torralba_2008_80}.
There are $50\rm{,}000$ images for training, and $10\rm{,}000$ images for testing, all of them are $32\times32$ RGB images.
{\bf CIFAR10} contains $10$ basic categories, and both training and testing data are uniformly distributed over these categories.
To avoid using the testing data, we leave $10\rm{,}000$ images from the training set for validation.

\subsubsection{Settings and Results}
\label{Experiments:CIFAR10:Results}

We follow a revised {\bf LeNet} for {\bf CIFAR10} recognition.
The original network is abbreviated as:
\begin{spverbatim}
C5(P2)@8-MP3(S2)-C5(P2)@16-MP3(S2)-
C5(P2)@32-MP3(S2)-FC128-D0.5-FC10.
\end{spverbatim}
\noindent
Note that we significantly reduce the filter numbers at each stage to accelerate the training phase.
We will show later that this does not prevent the genetic process from learning promising network structures.
We apply $120$ training epochs with learning rate $10^{-2}$, followed by $60$ epochs with learning rate $10^{-3}$,
$40$ epoch with learning rate $10^{-4}$ and another $20$ epoch with learning rate $10^{-5}$.

We keep the fully-connected part of the above network unchanged,
and set ${S}={3}$ and ${\left(K_1,K_2,K_3\right)}={\left(3,4,5\right)}$.
Similarly, the first convolutional layer within each stage remains the same as in the original {\bf LeNet},
and other convolutional layers take the kernel size $3\times3$ and the same channel number.
The length $L$ of each binary string is $19$, which means that there are ${2^{19}={524\rm{,}288}}$ possible individuals.

We create an initial generation with ${N}={20}$ individuals, and run the genetic process for ${T}={50}$ rounds.
Other parameters are set to be ${p_\mathrm{M}}={0.8}$, ${q_\mathrm{M}}={0.05}$, ${p_\mathrm{C}}={0.2}$ and ${q_\mathrm{C}}={0.2}$.
The mutation and crossover parameters $q_\mathrm{M}$ and $q_\mathrm{C}$ are set to be smaller because the strings become longer.
The maximal number of explored individuals is ${20\times\left(50+1\right)}={1\rm{,}020}\ll{524\rm{,}288}$.
The training phase of each individual takes an average of $0.4$ hour, and the entire genetic process takes about $17$ GPU-days.

\begin{table*}
\centering
\begin{tabular}{|l||r|r|r|r|r|c|}
\hline
Gen & Max $\%$         & Min $\%$         & Avg $\%$         & Med $\%$         & Std-D   & Best Network Structure            \\
\hline
00  & $75.96$          & $71.81$          & $74.39$          & $74.53$          & $ 0.91$ & {\tt 0-01|0-01-111|0-11-010-0111} \\
\hline
01  & $75.96$          & $73.93$          & $75.01$          & $75.17$          & $ 0.57$ & {\tt 0-01|0-01-111|0-11-010-0111} \\
\hline
02  & $75.96$          & $73.95$          & $75.32$          & $75.48$          & $ 0.57$ & {\tt 0-01|0-01-111|0-11-010-0111} \\
\hline
03  & $76.06$          & $73.47$          & $75.37$          & $75.62$          & $ 0.70$ & {\tt 1-01|0-01-111|0-11-010-0111} \\
\hline
05  & $76.24$          & $72.60$          & $75.32$          & $75.65$          & $ 0.89$ & {\tt 1-01|0-01-111|0-11-010-0011} \\
\hline
08  & $76.59$          & $74.75$          & $75.77$          & $75.86$          & $ 0.53$ & {\tt 1-01|0-01-111|0-11-010-1011} \\
\hline
10  & $76.72$          & $73.92$          & $75.68$          & $75.80$          & $ 0.88$ & {\tt 1-01|0-01-110|0-11-111-0001} \\
\hline
20  & $76.83$          & $74.91$          & $76.45$          & $76.79$          & $ 0.61$ & {\tt 1-01|1-01-110|0-11-111-0001} \\
\hline
30  & $76.95$          & $74.38$          & $76.42$          & $76.53$          & $ 0.46$ & {\tt 1-01|0-01-100|0-11-111-0001} \\
\hline
50  & $\mathbf{77.06}$ & $\mathbf{75.34}$ & $\mathbf{76.58}$ & $\mathbf{76.81}$ & $ 0.55$ & {\tt 1-01|0-01-100|0-11-101-0001} \\
\hline
\end{tabular}
\caption{
    Recognition accuracy ($\%$) on the {\bf CIFAR10} testing set.
    The zeroth generation is the initialized generation.
    We set ${S}={3}$ and ${\left(K_1,K_2,K_3\right)}={\left(3,4,5\right)}$.
}
\label{Tab:CIFAR}
\end{table*}

We perform two individual genetic processes.
The results of one process are summarized in Table~\ref{Tab:CIFAR}.
As in the {\bf MNIST} experiments, all the important statistics
({\em e.g.}, average and median accuracies) grow from generation to generation.
We also report the best network structures in the table,
and visualize the best structures throughout these two processes in Figure~\ref{Fig:Observation}.

\subsubsection{Comparison to Densely-Connected Nets}
\label{Experiments:CIFAR10:Comparison}

Under our encoding scheme, a straightforward way of network construction is to set all bits to be $1$,
which leads to a network in which any two layers within the same stage are connected.
This network produces a $76.84\%$ recognition rate, which is a little bit lower than those reported in Table~\ref{Tab:CIFAR}.
Considering that the densely-connected network requires heavier computational overheads,
we conclude that the genetic algorithm helps to find more effective and efficient structures than the dense connections.

\subsubsection{Observation}
\label{Experiments:CIFAR10:Observation}

\renewcommand{\figurewidth}{8.0cm}
\begin{figure}
\begin{center}
    \includegraphics[width=\figurewidth]{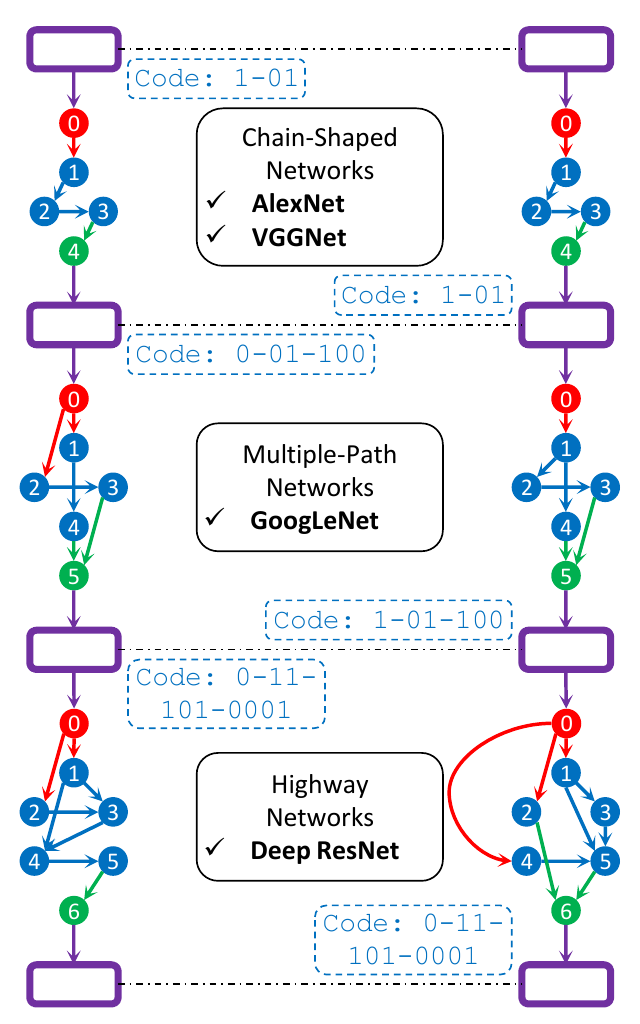}
\end{center}
\caption{
    Two network structures learned from the two independent genetic processes (best viewed in color PDF).
}
\label{Fig:Observation}
\end{figure}

In Figure~\ref{Fig:Observation}, we plot the the network structures learned from two individual genetic processes.
The structures learned by the genetic algorithm are quite different from the manually designed ones,
although some manually designed local structures are observed,
like the chain-shaped networks, multi-path networks and highway networks.
We emphasize that these two networks are obtained by independent genetic processes,
which demonstrates that our genetic process generally converges to similar network structures.

\subsection{CIFAR and SVHN Experiments}
\label{Experiments:CIFARSVHN}

We apply the networks learned from the {\bf CIFAR10} experiments to more small-scale datasets.
We test three datasets, {\em i.e.}, {\bf CIFAR10}, {\bf CIFAR100} and {\bf SVHN}.
{\bf CIFAR100} is an extension to {\bf CIFAR10} containing $100$ finer-grained categories.
It has the same numbers of training and testing images as {\bf CIFAR10},
and these images are also uniformly distributed over $100$ categories.

{\bf SVHN} (Street View House Numbers)~\cite{Netzer_2011_Reading} is a large collection of $32\times32$ RGB images,
{\em i.e.}, $73\rm{,}257$ training samples, $26\rm{,}032$ testing samples, and $531\rm{,}131$ extra training samples.
We preprocess the data as in the previous work~\cite{Netzer_2011_Reading},
{\em i.e.}, selecting $400$ samples per category from the training set as well as $200$ samples per category from the extra set,
using these $6\rm{,}000$ images for validation, and the remaining $598\rm{,}388$ images as training samples.
We also use Local Contrast Normalization (LCN) for preprocessing~\cite{Goodfellow_2013_Maxout}.

\newcommand{\colwidth}{1.0cm}
\begin{table}[t]
\begin{center}
\begin{tabular}{|l||R{\colwidth}|R{\colwidth}|R{\colwidth}|}
\hline
{}                                                   & {\bf SVHN}       & {\bf CF10}       & {\bf CF100}      \\
\hline\hline
Zeiler {\em et.al}~\cite{Zeiler_2013_Stochastic}     & $ 2.80$          & $15.13$          & $42.51$          \\
\hline
Goodfellow {\em et.al}~\cite{Goodfellow_2013_Maxout} & $ 2.47$          & $ 9.38$          & $38.57$          \\
\hline
Lin {\em et.al}~\cite{Lin_2014_Network}              & $ 2.35$          & $ 8.81$          & $35.68$          \\
\hline
Lee {\em et.al}~\cite{Lee_2015_Deeply}               & $ 1.92$          & $ 7.97$          & $34.57$          \\
\hline
Liang {\em et.al}~\cite{Liang_2015_Recurrent}        & $ 1.77$          & $ 7.09$          & $31.75$          \\
\hline
Lee {\em et.al}~\cite{Lee_2016_Generalizing}         & $ 1.69$          & $ 6.05$          & $32.37$          \\
\hline
Zagoruyko {\em et.al}~\cite{Zagoruyko_2016_Wide}     & $ 1.85$          & $ 5.37$          & $24.53$          \\
\hline
Xie {\em et.al}~\cite{Xie_2016_Geometric}            & $ 1.67$          & $ 5.31$          & $25.01$          \\
\hline
Huang {\em et.al}~\cite{Huang_2016_Deep}             & $ 1.75$          & $ 5.25$          & $24.98$          \\
\hline
Huang {\em et.al}~\cite{Huang_2016_Densely}          & $\mathbf{ 1.59}$ & $\mathbf{ 3.74}$ & $\mathbf{19.25}$ \\
\hline\hline
{\bf GeNet} after G-$00$                             & $ 2.25$          & $ 8.18$          & $31.46$          \\
\hline
{\bf GeNet} after G-$05$                             & $ 2.15$          & $ 7.67$          & $30.17$          \\
\hline
{\bf GeNet} after G-$20$                             & $ 2.05$          & $ 7.36$          & $29.63$          \\
\hline
{\bf GeNet} \#1 (G-$50$)                             & $ 1.99$          & $ 7.19$          & $\mathbf{29.03}$ \\
\hline
{\bf GeNet} \#2 (G-$50$)                             & $\mathbf{ 1.97}$ & $\mathbf{ 7.10}$ & $29.05$          \\
\hline
\end{tabular}
\caption{
    Comparison of the recognition error rate ($\%$) with the state-of-the-arts.
    We apply data augmentation on all these datasets.
    {\bf GeNet} \#1 and {\bf GeNet} \#2 are the structures shown in Figure~\ref{Fig:Observation}.
}
\label{Tab:SmallDatasets}
\end{center}
\end{table}

We evaluate the best network structure in each generation of the genetic process.
We resume using a large number of filters at each stage,
{\em i.e.}, the three stages and the first fully-connected layer
are equipped with $64$, $128$, $256$ and $1024$ filters, respectively.
The training strategy, include the numbers of epochs and learning rates, remains the same as in the previous experiments.

We compare our results with some state-of-the-art methods in Table~\ref{Tab:SmallDatasets}.
Among these competitors, we note that the densely-connected network~\cite{Huang_2016_Densely} is closely related to our work.
Although {\bf GeNet} ($17$ layers) produces lower recognition accuracy,
we note that the structures used
in~\cite{Zagoruyko_2016_Wide}\cite{Xie_2016_Geometric}\cite{Huang_2016_Deep}\cite{Huang_2016_Densely} are much deeper
({\em e.g.}, $40$--$100$ layers).
Since dense connection is often not the best option (see Section~\ref{Experiments:CIFAR10:Comparison}),
we believe that it is possible to use the genetic algorithm to optimize the connections used in~\cite{Huang_2016_Densely}.

\subsection{ILSVRC2012 Experiments}
\label{Experiments:ILSVRC2012}

We evaluate the top-$2$ networks on the {\bf ILSVRC2012} classification task~\cite{Russakovsky_2015_ImageNet},
a subset of the {\bf ImageNet} database~\cite{Deng_2009_ImageNet} which contains $1\rm{,}000$ object categories.
The training set, validation set and testing set contain $1.3\mathrm{M}$, $50\mathrm{K}$ and $150\mathrm{K}$ images, respectively.
The input images are of $224\times224\times3$ pixels.
We first apply the first two stages in the {\bf VGGNet} ($4$ convolutional layers and two pooling layers)
to change the data dimension to $56\times56\times128$.
Then, we apply the two networks shown in Figure~\ref{Fig:Observation},
and adjust the numbers of filters at three stages to $256$, $512$ and $512$ (following {\bf VGGNet}), respectively.
After these stages, we obtain a $7\times7\times512$ data cube.
We preserve the fully-connected layers in {\bf VGGNet} with the dropout rate $0.5$.
We apply the training strategy as in {\bf VGGNet}.
The entire training process of each network takes around $20$ GPU-days.

\begin{table}
\centering
\begin{tabular}{|l||r|r|r|}
\hline
{}                                            & Top-$1$ & Top-$5$ & Depth \\
\hline
{\bf AlexNet}~\cite{Krizhevsky_2012_ImageNet} & $42.6 $ & $19.6 $ & $8$   \\
\hline
{\bf GoogLeNet}~\cite{Szegedy_2015_Going}     & $34.2 $ & $12.9 $ & $22$  \\
\hline
{\bf VGGNet-16}~\cite{Simonyan_2015_Very}     & $28.5 $ & $ 9.9 $ & $16$  \\
\hline
{\bf VGGNet-19}~\cite{Simonyan_2015_Very}     & $28.7 $ & $ 9.9 $ & $19$  \\
\hline
{\bf ResNet-50}~\cite{He_2016_Deep}           & $24.6 $ & $ 7.7 $ & $50$  \\
\hline
{\bf ResNet-101}~\cite{He_2016_Deep}          & $23.4 $ & $ 7.0 $ & $101$ \\
\hline
{\bf ResNet-152}~\cite{He_2016_Deep}          & $23.0 $ & $ 6.7 $ & $152$ \\
\hline\hline
{\bf GeNet} \#1                               & $28.12$ & $ 9.95$ & $22$  \\
\hline
{\bf GeNet} \#2                               & $27.87$ & $ 9.74$ & $22$  \\
\hline
\end{tabular}
\caption{
    Top-$1$ and top-$5$ recognition error rates ($\%$) on the {\bf ILSVRC2012} dataset.
    For all competitors, we report the single-model performance without using any complicated data augmentation in {\em testing}.
    These numbers are copied from this page: {\tt http://www.vlfeat.org/matconvnet/pretrained/}.
    We use the networks shown in Figure~\ref{Fig:Observation}, and name them as {\bf GeNet} \#1 and \#2, respectively.
}
\label{Tab:ILSVRC2012}
\end{table}

Results are summarized in Table~\ref{Tab:ILSVRC2012}.
We can see that, in general, structures learned from a small dataset ({\bf CIFAR10})
can be transferred to large-scale visual recognition ({\bf ILSVRC2012}).
Our model achieves better performance than {\bf VGGNet},
because the original three chain-shaped stages are replaced by the automatically learned structures.

\section{Conclusions}
\label{Conclusions}

This paper applies the genetic algorithm to designing the structure of deep neural networks.
We first propose an encoding method to represent each network structure with a fixed-length binary string,
then uses some popular genetic operations such as mutation and crossover to explore the search space efficiently.
Different initialization strategies do not make much difference on the genetic process.
We conduct the genetic algorithm with a relatively small reference dataset ({\bf CIFAR10}),
and find that the generated structures transfer well to other tasks, including the large-scale {\bf ILSVRC2012} dataset.

Despite the interesting results we have obtained, our algorithm suffers from several drawbacks.
First, a large fraction of network structures are still unexplored,
including those with non-convolutional modules like Maxout~\cite{Goodfellow_2013_Maxout},
and the multi-scale strategy used in the inception module~\cite{Szegedy_2015_Going}.
Second, in the current work, the genetic algorithm is only used to explore the network structure,
whereas the network training process is performed separately.
It would be very interesting to incorporate the genetic algorithm to training the network structure and weights simultaneously.
These directions are left for future work.

\section*{Acknowledgements}
\label{Acknowledgements}

We thank John Flynn, Wei Shen, Chenxi Liu and Siyuan Qiao for instructive discussions.

{\small
\bibliographystyle{ieee}
\bibliography{egbib}
}

\end{document}